\documentclass[runningheads]{llncs}
\usepackage{graphicx}
\usepackage{comment}
\usepackage{amsmath,amssymb} 
\usepackage{color}
\usepackage{multirow}

\usepackage[width=122mm,left=12mm,paperwidth=146mm,height=193mm,top=12mm,paperheight=217mm]{geometry}

\newenvironment{packed_enum}{
\begin{enumerate}
  \setlength{\itemsep}{1pt}
  \setlength{\parskip}{0pt}
  \setlength{\parsep}{0pt}
}{\end{enumerate}}

\begin{document}
\pagestyle{headings}
\mainmatter

\title{MagicEyes: A Large Scale Eye Gaze Estimation Dataset for Mixed Reality} 

 \author{Zhengyang Wu\and Srivignesh Rajendran\and Tarrence van As \and Joelle Zimmermann \\ Vijay Badrinarayanan \and Andrew Rabinovich}
 \institute{Magic Leap Inc, USA}

\titlerunning{MagicEyes}
%
%
\authorrunning{Z. Wu et al.}
%
\maketitle

\begin{abstract}
With the emergence of Virtual and Mixed Reality (XR) devices, eye tracking has received significant attention in the computer vision community. Eye gaze estimation is a crucial component in XR -- enabling energy efficient rendering, multi-focal displays, and effective interaction with content. In head-mounted XR devices, the eyes are imaged off-axis to avoid blocking the field of view. This leads to increased challenges in inferring eye related quantities and simultaneously provides an opportunity to develop accurate and robust learning based approaches.
 To this end, we present MagicEyes, the first large scale eye dataset collected using real MR devices with comprehensive ground truth labeling. MagicEyes includes $587$ subjects with $80,000$ images of human-labeled ground truth and over $800,000$ images with gaze target labels. We evaluate several state-of-the-art methods on MagicEyes and also propose a new multi-task EyeNet model designed for 
 detecting the cornea, glints and pupil along with eye segmentation in a single forward pass.

\end{abstract}

\section{Introduction}
Eye gaze estimation is a critical component for current and future generations of head-mounted devices (HMDs) for virtual and mixed reality (often grouped as XR).  It enables energy and bandwidth efficient rendering of content (foveated rendering~\cite{guenter2012foveated}) and drives multi-focal displays for more realistic rendering of content (minimizing accommodation vergence conflict~\cite{oculustalk}).


 \begin{figure}[htp]
  \centering
    \begin{tabular}{c|c}
        \includegraphics[height=3.5cm,keepaspectratio]{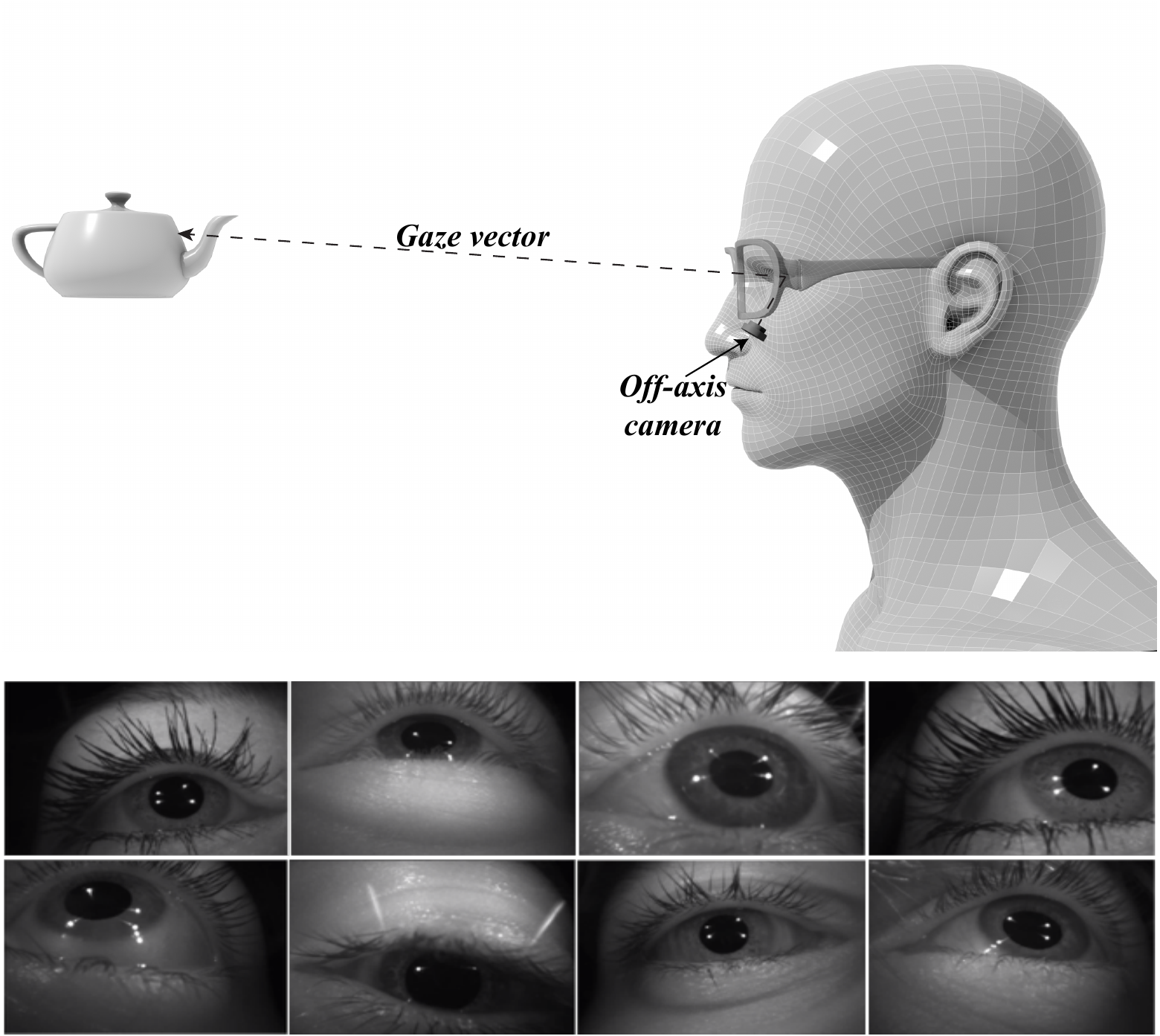} &
        \includegraphics[height=3.5cm,keepaspectratio]{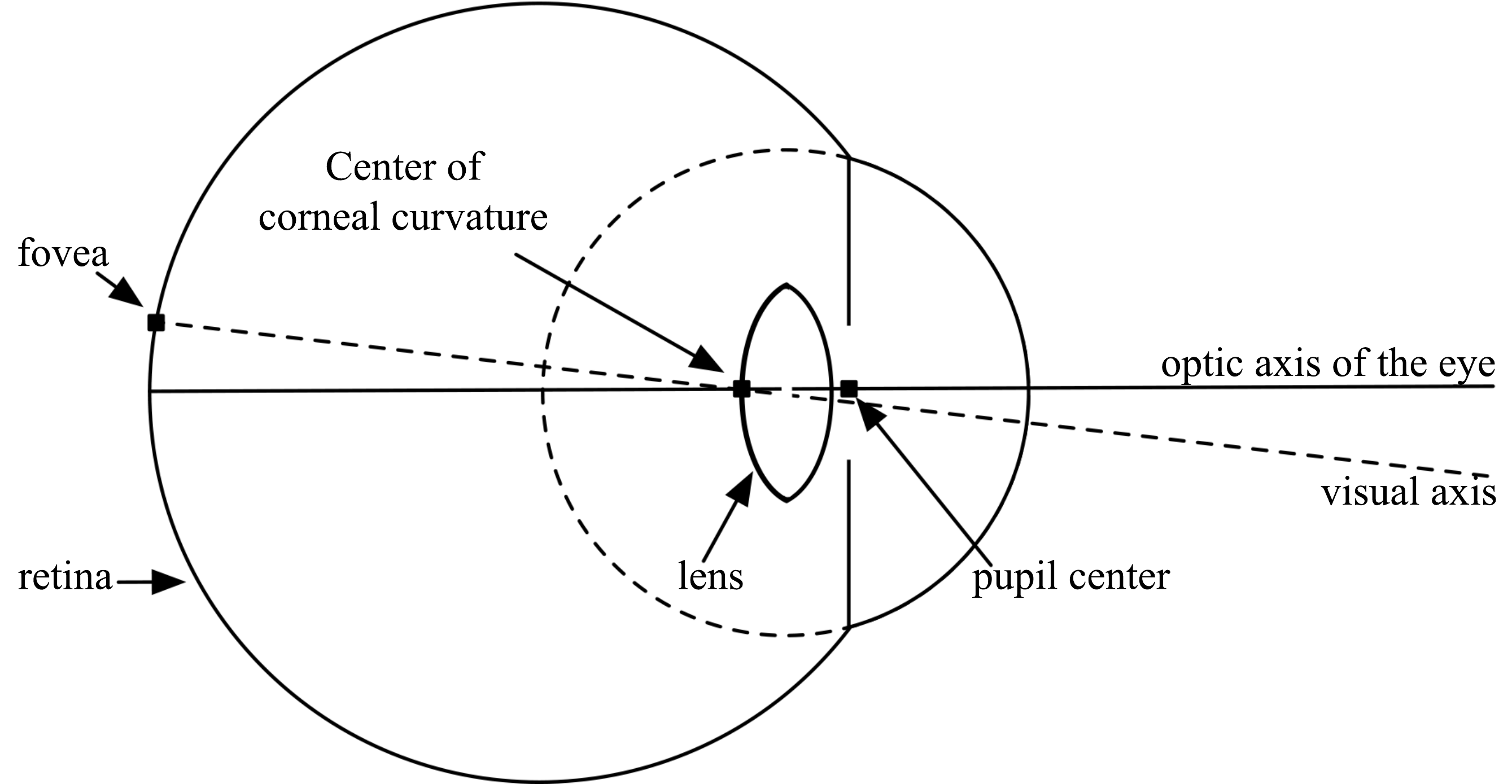} \\
        (a) & (b) \\
    \end{tabular}
    \caption[9pt]{(a) Off-axis IR camera setting in head-mounted VR/MR devices. Also shown are some sample eye images captured in this setting (b) An illustration of the standard geometric model of the physiology of human eye}
  \label{fig:illustration}
  \vspace{-3mm}
 \end{figure}

The typical setup in head-mounted Virtual and Mixed Reality devices for eye gaze estimation is illustrated in Figure~\ref{fig:illustration}(a). A set of four IR LEDs are placed in and around the display and their reflections (glints) are detected using the IR-sensitive eye camera, one for each eye. These glints are used to estimate important geometric quantities in the eye which are not directly observable from the eye camera images. Image samples from such a setup are shown in Figure~\ref{fig:illustration}(a). The four main semantic classes, Face/Background, Sclera, Iris and Pupil, along with the glints are clearly visible. As can be seen from these examples, there can be a large angle between the user's gaze and the camera axis. This makes eye gaze estimation challenging due to the increased eccentricity of pupils, partial occlusions caused by the eyelids and eyelashes~\cite{swirski2012robust}, as well as glint distractions caused due to environment illumination.
The challenging eye gaze estimation problem in such off-axis setting offers a great opportunity to research learning based approaches to both the appearance based (e.g. eye part segmentation) and geometric computations (cornea, pupil and gaze estimation) that are the key components of an eye tracking pipeline. We therefore introduce the MagicEyes dataset to benchmark off-axis eye gaze estimation pipelines on both appearance and geometric tasks. The dataset is the largest to date with $587$ subjects and over $800K$ gaze target labels covering a wide variety of age, ethnicity, eye and skin color, make up. We also provide accurate ground truth for important related problems such as eye part segmentation, glint detection. With this elaborate data collection and labeling effort, we hope to stimulate learning based eye gaze estimation research for XR devices in real life settings.


A standard geometric model of the human eye is shown in Figure~\ref{fig:illustration}(b). The eye ball sphere encompasses the inner corneal sphere and within the corneal sphere lies the pupil opening. The \textit{optical axis} of the eye connects the cornea center and the pupil (opening) center. The \textit{visual axis} or \textit{gaze} vector, for all practical purposes, is taken to be the line joining the cornea center and the fovea at the back of the eye. The angle ($\kappa$) between the gaze vector and optical axis is assumed to be constant for each user. Estimating the optical axis (pupil and cornea center) is the key problem underlying gaze tracking. Per subject calibration can be performed to transform the optical axis to the gaze vector.

Traditional gaze estimation pipelines use computer vision techniques involving heuristics to estimate eye features like the pupil center, glint like blobs, eye part segments. This is followed by a geometric model based computation to detect glints and estimate the cornea center (origin) for gaze estimation. In contrast, we show that by training our proposed multi-task EyeNet network appropriately we eliminate the need for heuristic methods to detect image features. Furthermore, EyeNet can simultaneously estimate robust values for glint positions, labels and the cornea center in a feed forward manner. The shared representation across several tasks enables robust estimation of all the desired quantities for gaze estimation while also amortizing the computational load.
\begin{figure*}[h!]
  \includegraphics[width=\textwidth,height=\textheight,keepaspectratio]{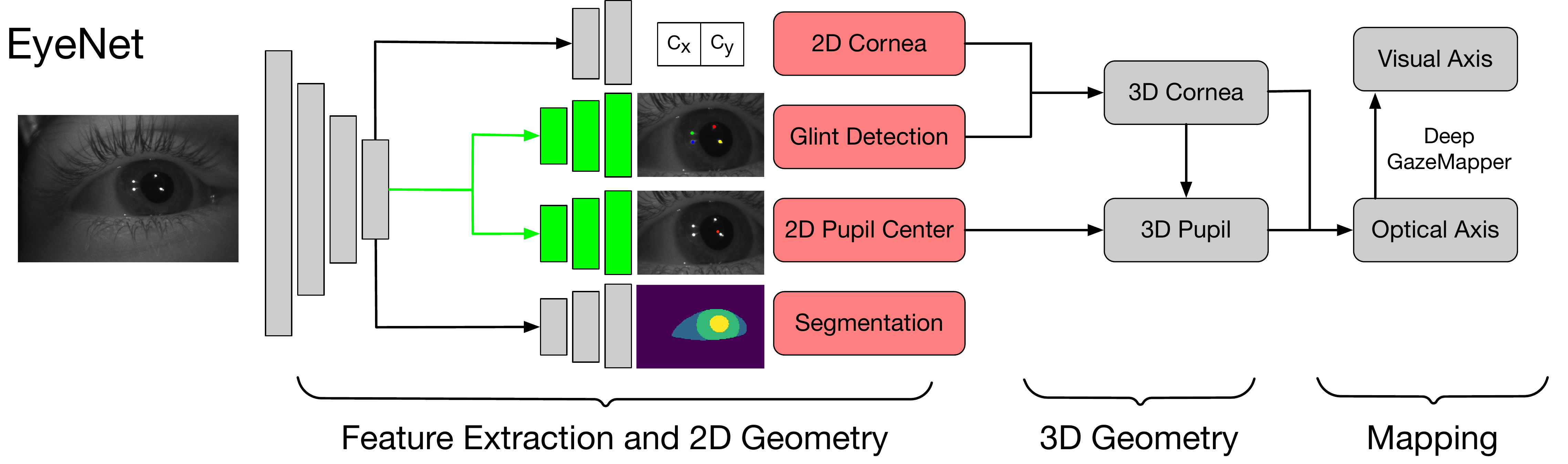}
  \caption[9pt]{EyeNet architecture where green boxes indicate weight sharing. Red boxes are the outputs of EyeNet with a single feature encoder network and four task decoders. Grey boxes correspond to 3D geometry and learnt gaze estimation using the DeepGazeMapper (This figure is best viewed in color)}
  \label{fig:pipeline}
  \vspace{-3mm}
\end{figure*}



To summarize, our contributions in this paper are:
\vspace*{-2mm}
\begin{packed_enum}
\item MagicEyes: first large-scale eye tracking related dataset collected using real mixed reality devices and labeled with comprehensive ground-truth
\item An evaluation of recent related work in eye segmenatic segmentation, pupil localization and gaze estimation on the MagicEyes dataset. By showing the pros and cons of each work, we aim to inspire the direction of future research.
\item EyeNet: The first multi-task deep neural network trained to jointly estimate multiple quantities relating to eye gaze estimation from off-axis eye images using both supervised learning and model based learning.
\end{packed_enum}

\vspace{-3mm}
\section{Related Work}
\subsection{Datasets}
There have been several gaze related datasets in recent years, well known public datasets include, Eye Chimera~\cite{florea2013can} consists of RGB images of 40 subjects at $1920\times1080$ resolution with manual markers; Columbia Gaze~\cite{CAVE_0324} includes 5880 head images of 56 subjects with $320\times240$ eye regions; Świrski and Dodgson~\cite{Swirski2013} created 158 synthetic, near-eye IR passive illumination images at $640\times480$ resolution; EYEDIAP~\cite{FunesMora_ETRA_2014} provides 16 subjects with eye images 192×168 resolution; UT Multi-view~\cite{sugano2014learning} has 64k near-eye images of 50 subjects and 1.1M synthetic images, both at $60\times36$ resolution; SynthesEyes~\cite{wood2015rendering} has 11.4k synthetic near-eye RGB images with passive illumination at $120\times80$ resolution; GazeCapture~\cite{krafka2016eye} crowd-sourced 2.5M mobile phone images from 1474 subjects; LPW~\cite{tonsen2016labelled} has 131k near-eye IR images with active illumination of 22 subjects at $640\times480$ resolution; MPIIGaze~\cite{zhang2017mpiigaze} consists of 214k webcam images of 15 subjects with $60\times36$ eye images; PupilNet 2.0~\cite{fuhl2017pupilnet} has 135k IR near-eye images with $384\times288$ eyes in varying lighting conditions; InvisibleEye~\cite{tonsen2017invisibleeye} has 280k images of 17 subjects from four 5×5 pixel cameras; WebGazer~\cite{papoutsaki2018eye} webcam video of 51 subjects with eye images at $640\times480$ resolution. NVGaze~\cite{kim2019nvgaze} has 2.5M images for 35 subjects.

All of these datasets fall into two categories, either synthetically generated eye images with noise-free gaze targets to mimic real-world settings, or collected using a camera (webcam) by displaying a target on a 2D screen. None of the datasets were collected using currently used headmounted XR devices in a real environment. Another key issue, particularly for emerging learning based methods, is the lack of large scale high quality datasets. Some datasets provide large amount of data in low resolutions while the other datasets include high quality data with small scale and limited diversity. In this work, we introduce a large scale dataset of real eye images, MagicEyes, captured with multiple XR devices in real world usage conditions. Importantly, MagicEyes has sufficient diversity in ethnicity, eye and skin color, with ground truth for several eye related inferences. We believe that this dataset will foster the development of learning based approaches for accurate eye gaze estimation.

\subsection{Methods}
The problem of eye gaze or point of regard (PoR) estimation~\cite{guestrin2006general} is most commonly studied in the context of two application scenarios. The first is for monitoring user attention or saliency for content that is projected in modern electronic devices such as laptops or phones~\cite{krafka2016eye,kimconvolutional,huang2018predicting}. These devices are held at a fixed (often known) distance from the user, and are able to image the whole face with good resolution. The task then is narrowed down to estimating which part of the screen (x-y coordinates) the gaze is directed towards. Recent work has shown successful application of CNNs for this task~\cite{krafka2016eye,kimconvolutional}. ~\cite{krafka2016eye} in particular proposes a CNN trained on a large dataset of 2.5M frames collected from $\sim$1500 subjects, and demonstrates a tracking error of $1-2cm$. In general, CNNs are being successfully employed for saliency detection in images~\cite{kruthiventi2015deepfix,CAT2000,Judd_2012,huang2018predicting}. The common theme behind all these approaches is that the estimate is a 2D position or saliency map, and all of these rely on \textit{appearance-only} eye gaze estimation.

A second and different line of approach to eye gaze estimation is to infer geometric quantities of the eye such as pupil and cornea centers, and use them to estimate the optical/visual axis (gaze) in 3D~\cite{guestrin2006general}. These methods typically rely on appearance based features (pupil and iris boundaries, glints, etc) to estimate geometric quantities, based on assumptions from a geometric eye model~\cite{villanueva2008geometry}. The overall aim of these eye tracking systems is to estimate the 3D gaze vector and the vergence/fixation/point of regard in 3D without recourse to a screen. Another important difference with respect to the previous class of methods is that such systems~\cite{tobiieyetracker} do not image the entire face, and instead have separate cameras to image each eye (similar to the setup in Figure~\ref{fig:illustration}(a)). Having individual low resolution eye cameras that are off the visual axis makes the problem much more challenging (increased eccentricity of the pupil/iris boundaries, occlusions due to eye-lids and eye-lashes). For such approaches with an off-axis camera setup data driven end-to-end solutions seem to suffer from low precision~\cite{shrivastava2017learning}.

Multiple approaches have been proposed to overcome the challenges posed by the off-axis eye tracking setup. Recent analysis-by-synthesis approach~\cite{wood2014eyetab} uses large-scale procedurally rendered synthetic data for nearest neighbor based gaze estimation. While the results look encouraging, the precision is not ideal. Shape based methods~\cite{tong2010unified,hansen2005eye} use pupil and iris boundary shape to compute gaze. The use of traditional computer vision techniques to estimate region boundaries makes them fragile, with failures frequently occurring under lighting variations and low image quality.  The Corneal reflection methods use external light sources (LED) to create a pattern of glints on each eye~\cite{zhu2007novel,guestrin2006general,villanueva2008geometry,yoo2005novel,hennessey2006single}. The location of the LEDs and their corresponding glint reflections in the eye are used to estimate the position of cornea in 3D. This along with pupil position estimation provides the optical axis of the camera. These methods tend to have the highest precision in gaze estimation but rely on traditional image processing to estimate pupil position and glint locations, which compromises their robustness.

A recent work~\cite{kim2019nvgaze} aims to train deep neural networks for low latency gaze estimation using largely synthetically generated eye images. The results show the promise of using synthetic data at large scale to perform end to end learning when close attention is paid to eye modeling and rendering images according to real camera characteristics. EyeNet itself could benefit from training on such synthetic data when available. However, their experiments were restricted to on-axis images with real data collected in controlled settings which reduce noise in the ground truth. In practice, data collected in the wild for any user is usually noisy. Unlike their experiments where target is displayed on a screen at a fixed distance, real calibration targets are virtual objects in complex environments. When users are prompted to focus on them, invariably there exist errors due to loss of concentration and distraction, and hence the quality of the ground truth gaze is noisy. In comparison, our experiments were conducted on a much larger scale and in realistic settings representative of those found in the wild. Even when noisy ground truth is used for training, our results are competitive with state of the art trackers while simultaneously providing a variety of useful inferences.

\vspace{-3mm}
\section{EyeNet}
\vspace{-3mm}
Gaze estimation has been shown to be a challenging learning task for deep networks. Recent end-to-end learning approaches~\cite{krafka2016eye,shrivastava2017learning} have achieved relatively low gaze tracking precision. While another recent work~\cite{kim2019nvgaze} has sought to overcome this with careful synthetic eye renderings it has mainly done so for an on-axis camera setting and relatively noise free real data. Low precision for end-to-end approaches in real world settings can be attributed to the fact that direct gaze regression completely ignores the eye geometry which is a valuable prior. Inspired by the recent work in SLAM and scene understanding research where integrating geometry has proved to help deep nets get better at depth estimation~\cite{garg2016unsupervised} and relative pose prediction~\cite{handa2016gvnn,zhou2017unsupervised}, we propose a multi-task deep network that estimates intermediate quantities that geometrically relate to gaze estimation.


Fig.~\ref{fig:pipeline} shows our overall network structure. It consists of a feature encoding base network and four task branches for eye parts semantic segmentation, pupil center estimation and glint localization, pupil and glints presence classification, and 2D cornea estimation. We describe each component in detail below.

\vspace{-3mm}
\subsection{Feature Encoding Layers}
\label{sec:fel}
The feature encoding layers serve as the backbone of EyeNet, and they are shared across every task. We employ ResNet50~\cite{he2016deep} with a feature pyramid (FPN)~\cite{lin2017feature} to capture information across different scales. The eye image is $160\times120$ and features from the topmost encoder layer ($20\times15\times256$) are input to the task branches. Architectural details of EyeNet are provided in the supplementary.

\vspace{-3mm}
\subsection{Eye Parts Segmentation}
\label{sec:eye_parts_segmentation}
Eye parts segmentation consists four segmentation classes: Background, Sclera, Iris and Pupil. For this task, we take the last layer feature map from the encoder network and up-sample it using deconvolutional layers to the same resolution as the input image, similar to~\cite{badrinarayanan2017segnet,ronneberger2015u}. The resulting four channel output is converted to class probabilities using a softmax layer for each pixel independently. We then use a conventional cross-entropy loss between the predicted probability distribution and the one-hot labels obtained from manually labeled ground truth.

\vspace{-3mm}
\subsection{Pupil and Glint Localization}
\label{sec:pupil_glints_detect}
The pupil and glint localization branch gives us the pixel location of the four glints and pupil center, (total of five keypoints). The network decoder layers for these two tasks are similar to the eye parts segmentation branch and predict a set of five dense maps at the output corresponding to the five keypoints. Each dense map is (channel-wise) normalized to sum to unity across all the pixels. A cross-entropy loss is then calculated across all the pixels of each map during training. At test time, the center of the pupil or a particular glint is the pixel location corresponding to maximum probability.

In realistic settings, glints and/or the pupil center can often be occluded by the closing of eyelids, nuisance reflections can appear as glints and for some gaze angles glints may not appear on the reflective corneal surface. Therefore it is important to learn to  robustly classify the presence or absence of glints and the pupil center. For this task, we take the topmost layer feature map from the EyeNet encoder, use one convolution layer to reduce the number of feature channels and add one trainable fully-connected layer to produce a 5$\times$2 sized output. Each pair represents the presence, absence probability for one of the four glints and/or the pupil center. We use a binary cross-entropy loss to learn from human labeled ground truth.

\begin{figure}[htp]
  \centering
    \begin{tabular}{c|c}
        \includegraphics[height=3.5cm,keepaspectratio]{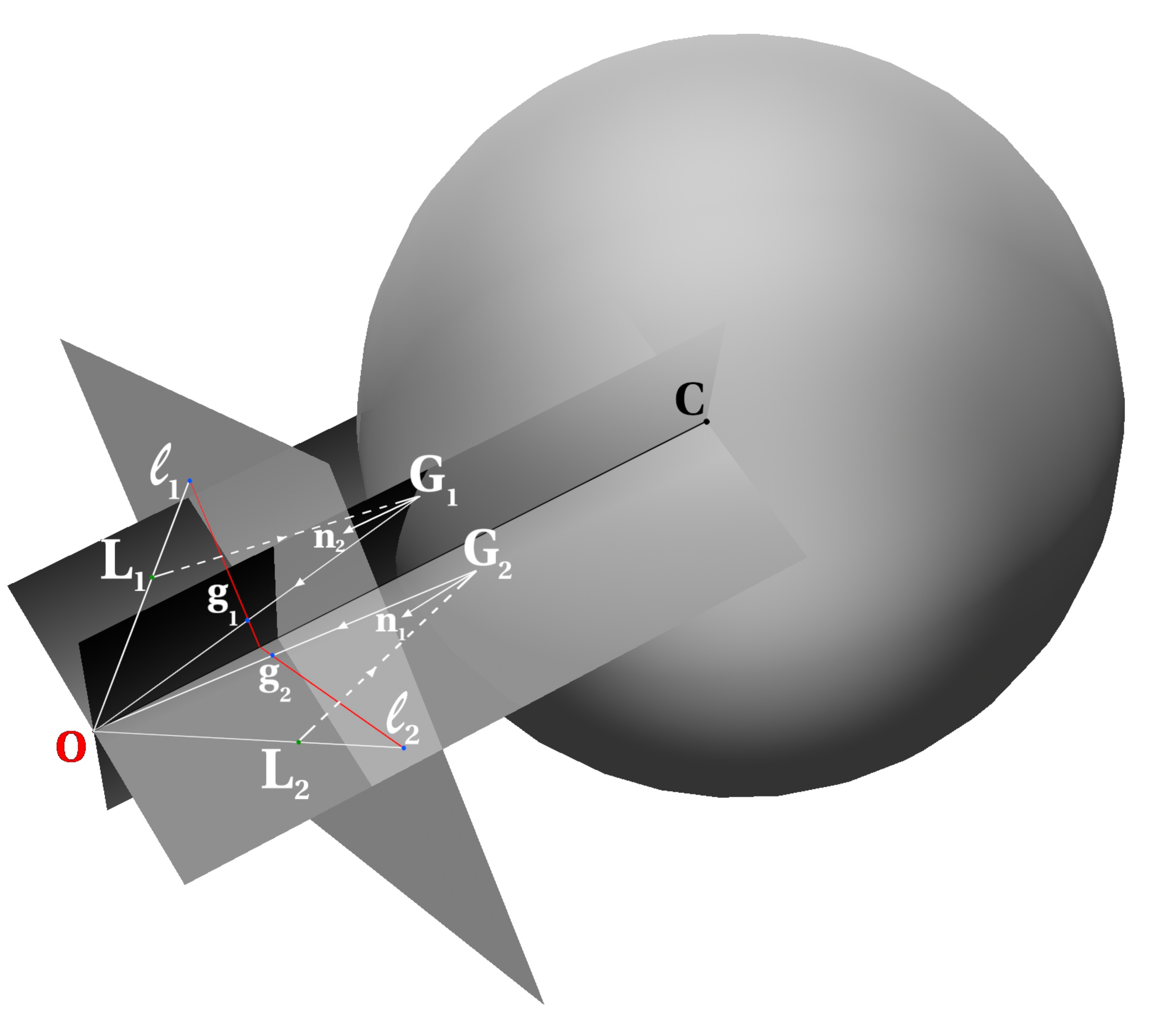} &
        \includegraphics[height=3.5cm,keepaspectratio]{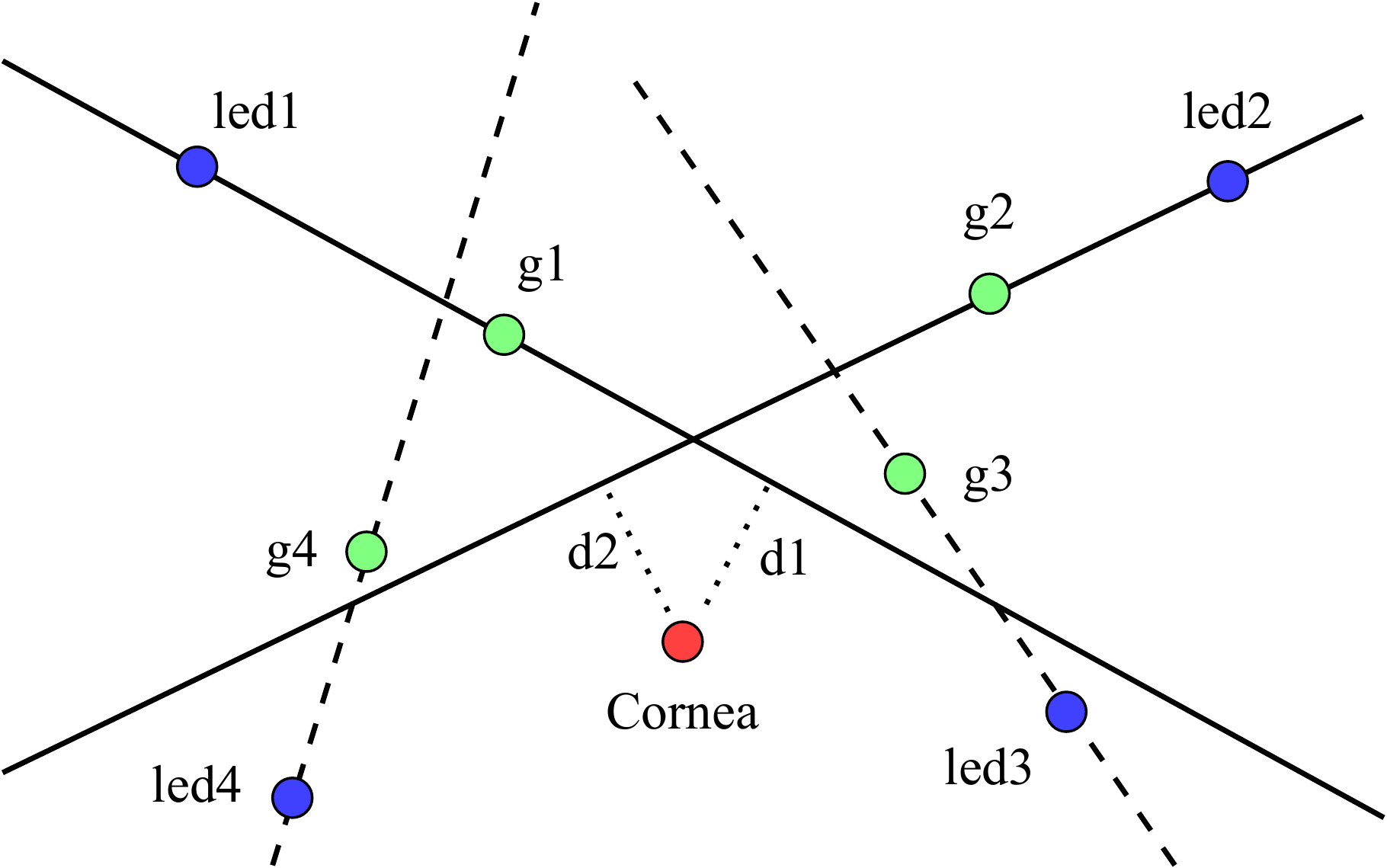} \\
        (a) & (b) \\
    \end{tabular}
    \caption{(a) An illustration of the geometric setup used to derive the model based supervision for cornea center training. The corneal sphere along with the camera image plane are shown. According to the law of light reflection, the incident ray $L_{1}G_{1}$, the reflected ray $G_{1}O$ and the normal $n_{1}$ at the point of incidence/reflection are co-planar. The same holds for the other LED. By this constraint, the cornea lies on the ray $OC$ intersecting the two planes. (b) An illustration of the standard geometric model of the physiology of human eye. (b)Projections of the LED, cornea 3D location onto the image plane shown along with the glint positions. The cornea 2D position should ideally lie on the intersection of all four LED-glint lines. We optimize cornea 2D to minimize the average distance from cornea 2D position to all the glint-led lines}
  \label{fig:device2}
  \vspace{-3mm}
 \end{figure}


\vspace{-3mm}
\subsection{Model Based Learning for Cornea Center Estimation}
\label{sec:geometry}
The center of the cornea is a geometric quantity in 3D which cannot be observed in a 2D image of an eye. Hence, unlike pupil (center of pupil ellipse) or glint labeling, it is not possible to directly hand label the projected location of the 3D cornea center on the image. We therefore propose a novel model-based learning method to predict the 2D cornea center.

Predicting the cornea using the learned network has two main benefits over using geometric constraints during evaluation (i) it's more robust because deep networks have a tendency to average out noise during training and standard out-of-network optimization can occasionally yield no convergence, and (ii) it only incurs a small and constant time feed forward compute since the cornea task branch consists of only a few fully connected layers.

Figure~\ref{fig:device2}(a) shows the geometric setup used to derive the model based supervision for training the cornea branch. The corneal sphere, two sample LED locations $L_{1},L_{2}$ in 3D, the camera center $O$ and the image plane are shown. The incident rays $L_{1}G_{1}$ and $L_{2}G_{2}$ from the LEDs and the corresponding reflected rays $G_{1}O$, $G_{2}O$ can be seen along with the normals $n_{1},n_{2}$ at the points of incidence/reflection. Given the cornea is a reflective surface it follows from the laws of light reflection an incident ray from a point source, e.g. $L_{1}G_{1}$, reflected ray $G_{1}O$ and the normal $n_{1}$ at the point of incidence/reflection on the sphere $G_{1}$ are all co-planar. Equivalently the corneal center $C$, LED $L_{1}$, the projection of the incident point $G_{1}$ onto the image plane denoted $g_{1}$ all lie on the same plane. This holds for other LED point sources too. More details can also be found in~\cite{villanueva2008geometry}.


From Figure~\ref{fig:device2}(a) we note that the line joining the projection of an LED onto the image plane $l_{i}$ and its corresponding glint location $g_{i}$ on the image plane intersects the cornea ray $OC$ due to the co-planarity constraints discussed above. A pair or more of such lines will intersect at the cornea center projection (cornea 2D) on the 2D image plane. The image plane view of this model based argument is shown in Figure~\ref{fig:device2}(b). The loss that we employ to train the cornea 2D branch is minimizing the distance from cornea 2D to all $l_{i},g_{i}$ lines as shown in Figure~\ref{fig:device2}(b).


Given the physiology of an individual's eyes is unique, we find it necessary to train the cornea branch to each subject (personalization) in order to produce the most accurate results, sharing the same idea with recent eye tracking works~\cite{zhang2018training,linden2018appearance}. Estimating 3D cornea ceter from 2D is described in the next section.

\vspace{-3mm}
\subsection{Lifting 2D Cornea Center to 3D}
From the cornea geometry in Figure~\ref{fig:device2}(a) at any hypothesized cornea 3D location, we can compute the ray passing through a 2D glint e.g. $g_{1}$ and its intersection with the corneal sphere (of assumed radius $r$). This ray can be reflected about the normal at the point of intersection $G_{1}$. If the hypothesized corneal 3D location is correct this reflected ray should pass very close to the known LED location $L_{1}$. We therefore minimize this distance for all the visible glints to estimate the best cornea 3D position. Note that we initialize the cornea 3D location with a typical distance between cornea and the camera center and then employ a standard gradient descent to minimize the loss. In practice, this out-of-network optimization takes fewer than 100 iterations to reach convergence.

\vspace{-3mm}
\subsection{Gaze Estimation}
\label{sec:gaze_estimation}
Following the illustration in Figure~\ref{fig:illustration}(b), given the estimate of the pupil center on the image plane and the cornea center in 3D, we can connect the camera center and pupil 2D to derive the pupil normal direction along which direction the pupil 3D center lies. We extend the pupil normal to intersect with the corneal ball to get an approximate pupil center in 3D. By connecting pupil 3D and cornea center 3D, we get an estimate of the optical axis.

We learn to estimate the gaze direction or equivalently the visual axis by training a gaze mapping network termed as the DeepGazeMapper. This deep network takes as input the optical axis direction and outputs the visual axis direction. Specifically, the visual axis is defined as the line connecting cornea 3D center to the calibrated visual target location.
In practice, we transform the optical axis and visual axis to the device coordinate system and unit normalize them before training. When the cornea 3D is also transformed to this coordinate system the gaze vector can be drawn through the cornea center and displayed.

In existing methods, a mapping function in the form of a hand prescribed second order polynomial function is fit to map pupil position to the 3D target position, effectively producing
the visual axis~\cite{guestrin2006general}. It is also possible to use a second order polynomial function to map the optical axis to the visual axis as is done in our implementation of the classical pipeline. This function is learnt for each subject using the calibration frames. We have shown here that this function can be replaced with a simple fully connected deep network (5 layers and ~30K params) also trained on only the calibration frames, described in Section~\ref{sec:Exp}.

\section{MagicEyes Dataset}
MagicEyes is the first large scale eye dataset collected on real mixed reality devices. Scale, diversity and similarity to real-world environments are all crucial factors for training an effective model that can generalize to practical use in the wild. In this section, we will describe the detailed statistics of MagicEyes, the device and procedures used to collect the dataset, and the kinds of ground truth data that's available to support gaze estimation and eye-tracking related tasks.

\vspace{-3mm}
\subsection{Statistics}
MagicEyes includes raw left and right eye images of 587 subjects across different genders, demographic groups with diverse physiology such as skin and eye colors. Table~\ref{tab:dataset}. shows the distribution of subjects across various characteristics such as gender, age group, ethnicity, eye color, skin color and presence of mascara. Such diverse characteristics helps researchers to conduct meaningful experiments that can reflect real-world use cases. The table also shows detailed train and test dataset splits used in the experiments. In order to test cross-device generalization capability, MagicEyes is collected using two devices with same components but different camera configurations. We used the first device (\textbf{D1}) to capture data of \textbf{427} subjects and the second device (\textbf{D2}) for \textbf{160} subjects. In total, there are over 3 million images collected over several months of effort.
\begin{table}[h!]
\small
\centering
\setlength{\tabcolsep}{0.5em}
\def\arraystretch{1}
\vspace{-3mm}
\caption{Demographic distribution for each dataset in MagicEyes. We try to make the distribution similar across all sets and mimic the proportions in the real world}
\begin{tabular}{|c|c|c|c|c|}
\hline
\textbf{Characteristics}    & \textbf{Types}  & \textbf{Train Set} & \textbf{Test Set D1} & \textbf{Test Set D2} \\ \hline
\multirow{2}{*}{Gender}     & Male            & 51.2               & 48.4                  & 46.7              \\ \cline{2-5}
                            & Female          & 48.8               & 51.6                  & 53.3              \\ \hline
\multirow{5}{*}{Age}        & 18-24           & 9.0                & 6.5                   & 15.6              \\ \cline{2-5}
                            & 25-34           & 36.2               & 31.2                  & 30.2              \\ \cline{2-5}
                            & 35-44           & 29.3               & 33.3                  & 26.9              \\ \cline{2-5}
                            & 45-54           & 19.5               & 20.4                  & 24.0              \\ \cline{2-5}
                            & 55-59           & 6.0                & 8.6                   & 3.3               \\ \hline
\multirow{4}{*}{Ethnicity}  & Caucasion       & 58.4               & 54.8                  & 59.0              \\ \cline{2-5}
                            & African Descent & 23.7               & 29.0                  & 17.0              \\ \cline{2-5}
                            & Hispanic        & 12.3               & 10.8                  & 15.1              \\ \cline{2-5}
                            & Other           & 5.6                & 5.4                   & 8.9               \\ \hline
\multirow{4}{*}{Eye Color}  & Brown           & 63.7               & 64.6                  & 53.3              \\ \cline{2-5}
                            & Blue            & 10.2               & 15.1                  & 19.8              \\ \cline{2-5}
                            & Hazel           & 17.1               & 14.0                  & 19.8              \\ \cline{2-5}
                            & Green           & 9.0                & 4.3                   & 7.1               \\ \hline
\multirow{4}{*}{Skin Color} & Light           & 57.8               & 51.6                  & 62.3              \\ \cline{2-5}
                            & Medium          & 15.0               & 17.2                  & 15.1              \\ \cline{2-5}
                            & Brown           & 8.1                & 6.3                   & 8.5               \\ \cline{2-5}
                            & Black           & 19.1               & 26.9                  & 14.1              \\ \hline
\multirow{2}{*}{Mascara}    & Yes             & 21.0               & 17.2                  & 22.2              \\ \cline{2-5}
                            & No              & 79.0               & 82.8                  & 77.8              \\ \hline
\end{tabular}
\label{tab:dataset}
\end{table}

\vspace{-3mm}

\subsection{Data Collection}
During the data collection phase, each subject is invited to a demo room mimicking the settings of a typical living room to help the subject relax and reduce the friction between the experiment setting and a real-world setting. After getting familiar with the environment, the subject is asked to put on a (Magic Leap One) mixed reality headset. Once the subject is comfortable with the headset and the headset is stable on the subject, we start to display points in an order on a virtual screen. The subject is asked to fixate his/her gaze on each gaze target point for two seconds and then click a button in his/her hands, after which a three seconds video of the subject's eye region is captured at 30fps (total of 90 images). Figure~\ref{fig:illustration}(a) shows some captured sample images of a subject's eye. The gaze targets are displayed in a $3\times3$ grid fashion on 6 depth planes simulating a distance from 0.5m, 0.75m, 1m, 1.5m, 2m, 3m respectively. We also alternate the space between each target (within the $3\times3$ grid) for every depth plane. We attempted to increase the density of the targets, but resulted in subjects failing to distinguish and focus on targets close to each other. Figure~\ref{fig:GTExample}(b) shows the virtual target grid we used to capture data. In total, there are 54 gaze targets used for every subject. For each subject, data collection takes about 10 minutes excluding setup and post-processing time.

\vspace{-3mm}
\subsection{Ground Truth}
MagicEyes includes ground truth gaze target position in 3D for every frames collected on both devices. In addition to gaze target locations, we also manually labeled eye segmentation, pupil and four glint locations. These can serve as intermediate supervision labels in hybrid pipelines using learning based estimation and geometric computation (e.g. EyeNet). Table~\ref{tab:gt} shows the ground truth availability for each quantity.



\begin{table}[]
\small
\centering
\setlength{\tabcolsep}{0.5em}
\def\arraystretch{1}
\vspace{-3mm}
\caption{Label availability for each ground truth type. Due to the difficulty of manual labeling of eye part segmentation, it is only available for a subset of all eye images collected. Even with this limitation, MagicEyes is still the largest dataset tailored for eye gaze estimation with manual labeling}
\begin{tabular}{|c|c|c|c|}
\hline
Data Type    & Train Set & Test Set Device 1 & Test Set Device 2 \\ \hline
Subjects     & 334       & 93                & 160               \\ \hline
Images       & 62K       & 18K               & 800K              \\ \hline
Segmentation & 62K       & 18K               & No                \\ \hline
Pupil Center & 62K       & 18K               & No                \\ \hline
Gaze Target  & 62K       & 18K               & 800K              \\ \hline
\end{tabular}
\label{tab:gt}
\end{table}

\begin{figure}[htp!]
  \centering
    \begin{tabular}{c|c}
        \includegraphics[height=3.5cm,keepaspectratio]{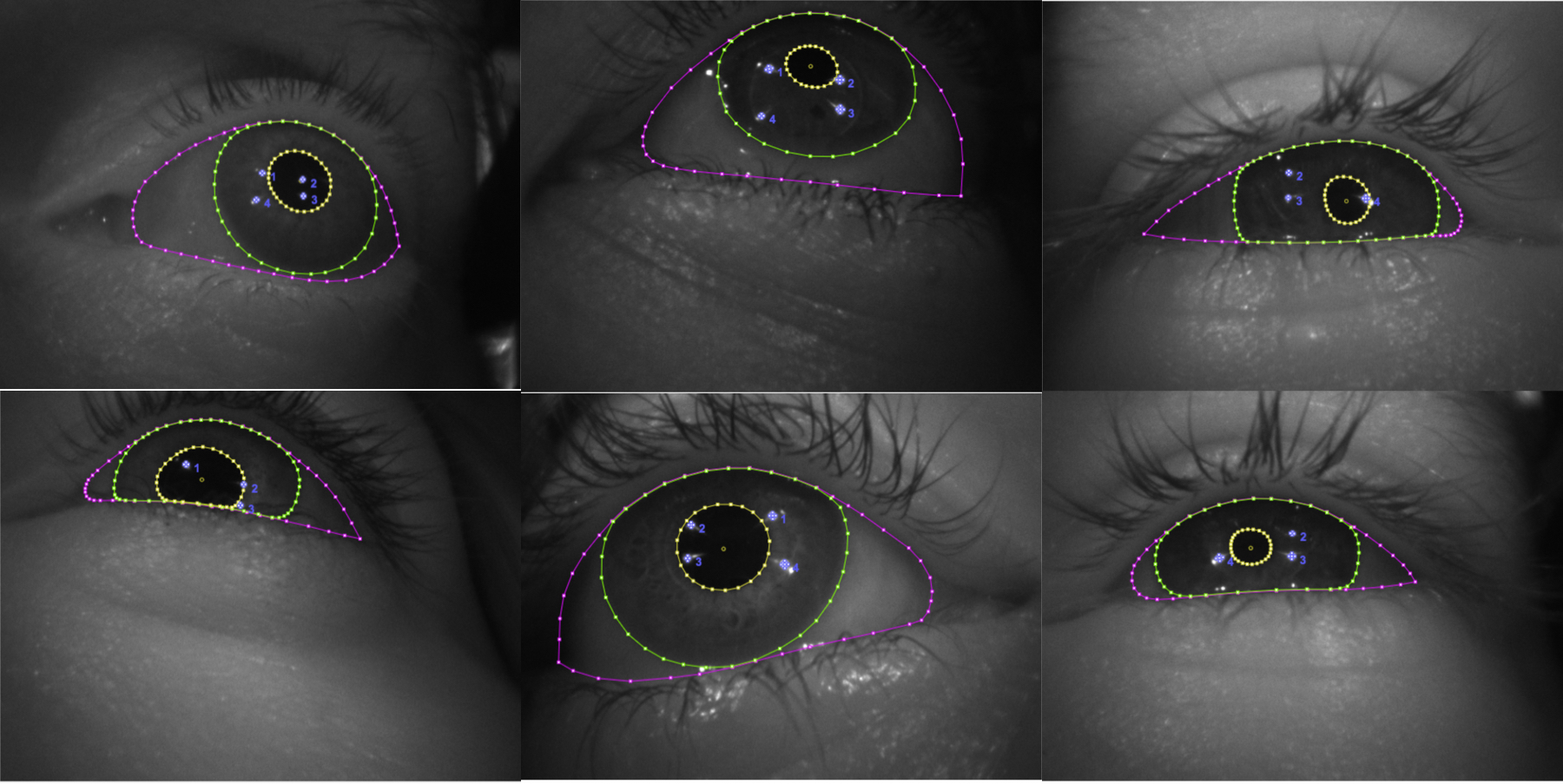} &
        \includegraphics[height=3.5cm,keepaspectratio]{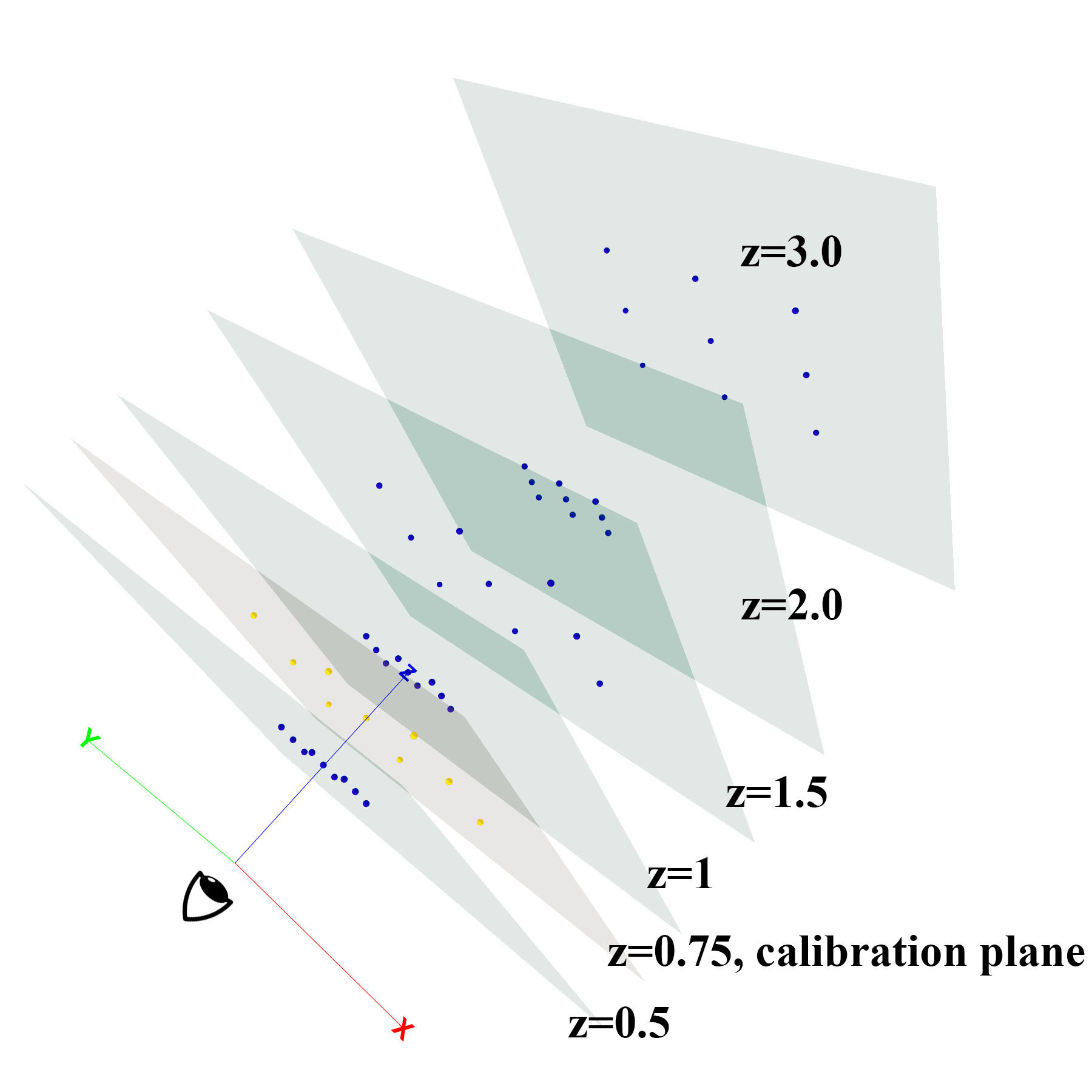} \\
        (a) & (b) \\
    \end{tabular}
    \caption[9pt]{(a) MagicEyes ground truth labels for segmentation and pupil localization. (b) Visualization of gaze targets on each of the 6 planes used in our experiments where Z is the distance from the user. Targets in second plane are used in personalizing models}
  \label{fig:GTExample}
  \vspace{-3mm}
 \end{figure}

We   presented   annotators   with   2D   eye   images   and
polygon marking, ellipse fitting tools. Each of the eye parts,
pupil, iris, sclera and background are manually annotated as shown in Figure~\ref{fig:GTExample}(a), by finely
marking the vertices of their corresponding polygon. The pupil center is then derived by fitting an ellipse to the pupil polygon vertices and noting its center. To annotate the glint
locations  and  labels  (association  with  the  corresponding
LED)  in  a  particular  frame,  human  labelers view past and
future  frames  to  mark  the  glint  blobs  and  also  note  the
presence or absence of a glint.  Each glint location is then
obtained  by  fitting  an  ellipse  to  the  marked  glint  blobs and estimating its center. The  glint  labels  are  assigned  by  observing  their  relative
positions. It takes on average 5 minutes to accurately label and review all ground truth. Metadata such as the camera intrinsics, LED position relative to headset are also included in MagicEyes.

\vspace{-3mm}
\subsection{Training and Testing Procedures For EyeNet}
The complete training takes several steps because our framework receives ground truth from different sources as well as geometric model-based supervision which uses bootstrapped estimates from the trained network itself. The first step is to train the ResNet~\cite{he2016deep} encoder-decoder network with eye segmentation labels because it provides the richest semantic information and is the most complicated supervised task to train accurately. Secondly, we use human labeled glint, pupil 2D center data \footnote{Obtained by fitting an ellipse to the human labelled pupil boundary.} and eye segmentation data together to jointly train all of these three supervised tasks. We find that initializing with weights trained from eye segmentation results in much more stable training than training from scratch. After this step, we generate glint predictions for all frames and use these along with known LED locations to train cornea 2D branch as discussed in Section~\ref{sec:geometry}.

We use the predicted cornea (personalized) and pupil 3D centers from the calibration frames to deduce the optical axis. Using the gaze target ground truth from calibration frames, we train the DeepGazeMapper to transform the optical axis to the visual axis. During test time, we obtain the predicted cornea and pupil 2D centers from EyeNet, lift  quantities to 3D and get the optical axis, which is then fed into the gaze mapping net to infer the predicted gaze direction.


\vspace{-3mm}
\section{Experiments}
\label{sec:Exp}
To present the usefulness of MagicEyes, we evaluated several state-of-the-art algorithms in various tasks related to gaze estimation. The classic way to predict gaze target is to first retrieve pupil and glint locations by either segmenting pupil boundary or direct inferring 2D pupil center, then estimate final gaze target based on inferred pupil and glint locations. Consequently, we evaluate three eye tracking related tasks including eye segmentation, pupil localization and then gaze estimation in this work.

\vspace{-3mm}
\subsection{Eye Segmentation}
Eye parts segmentation is defined as the task of assigning every pixel in the input image a class label from one of Background, Sclera, Iris and Pupil. We evaluated two state-of-the-art segmentation algorithms for this task. One is the winner of the recent OpenEDS workshop: RlTNet~\cite{chaudhary2019ritnet}, the other is the segmentation predictions from the proposed EyeNet. Since the input resolution of two methods are different, we resize the original dataset resolution ($640\times480$) to the input size of each model, then re-scale their predictions back to full resolution for a fair comparison.

\begin{figure}[h!]
  \centering
    \begin{tabular}{c|c}
        \includegraphics[height=3.5cm,keepaspectratio]{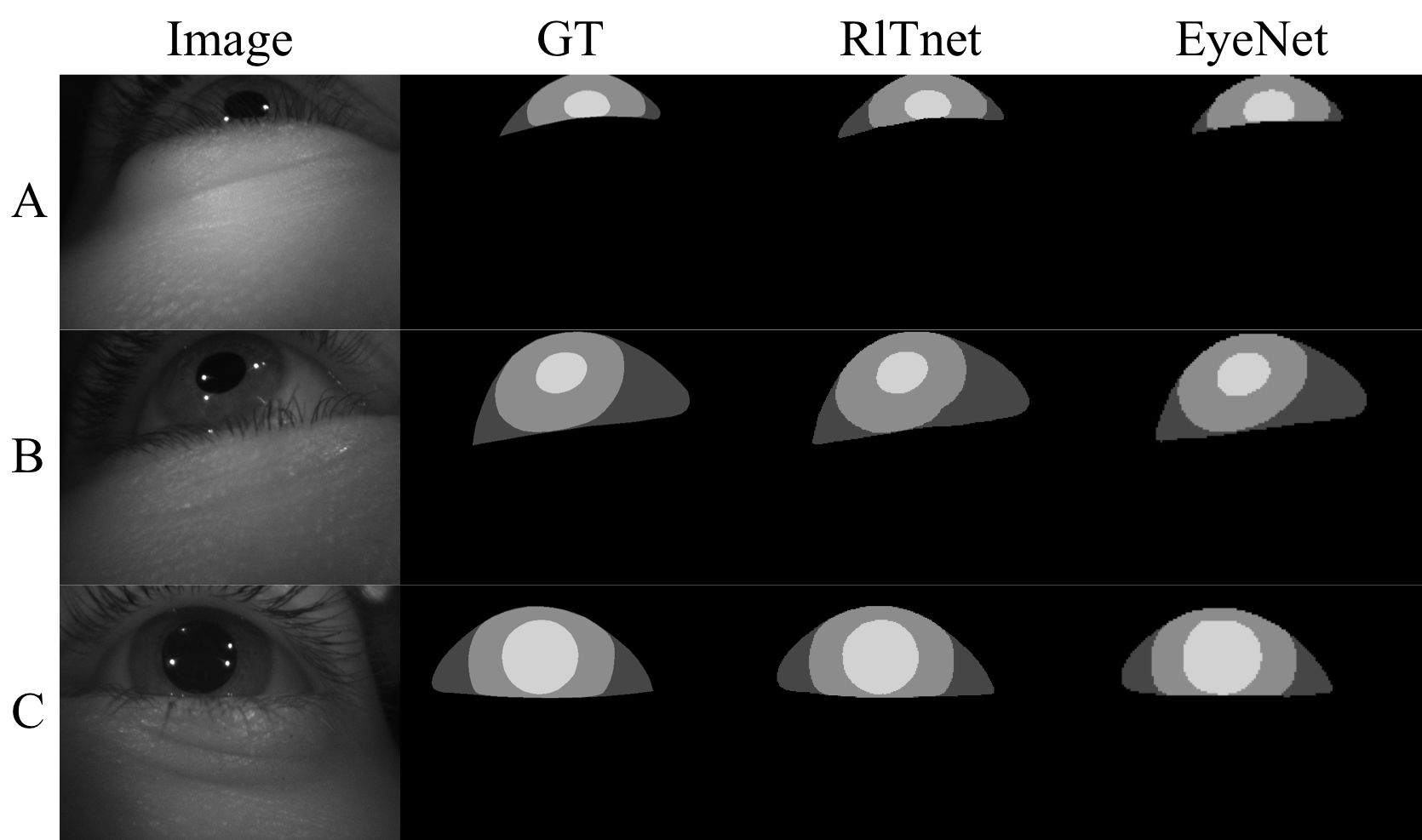} &
        \includegraphics[height=3.5cm,keepaspectratio]{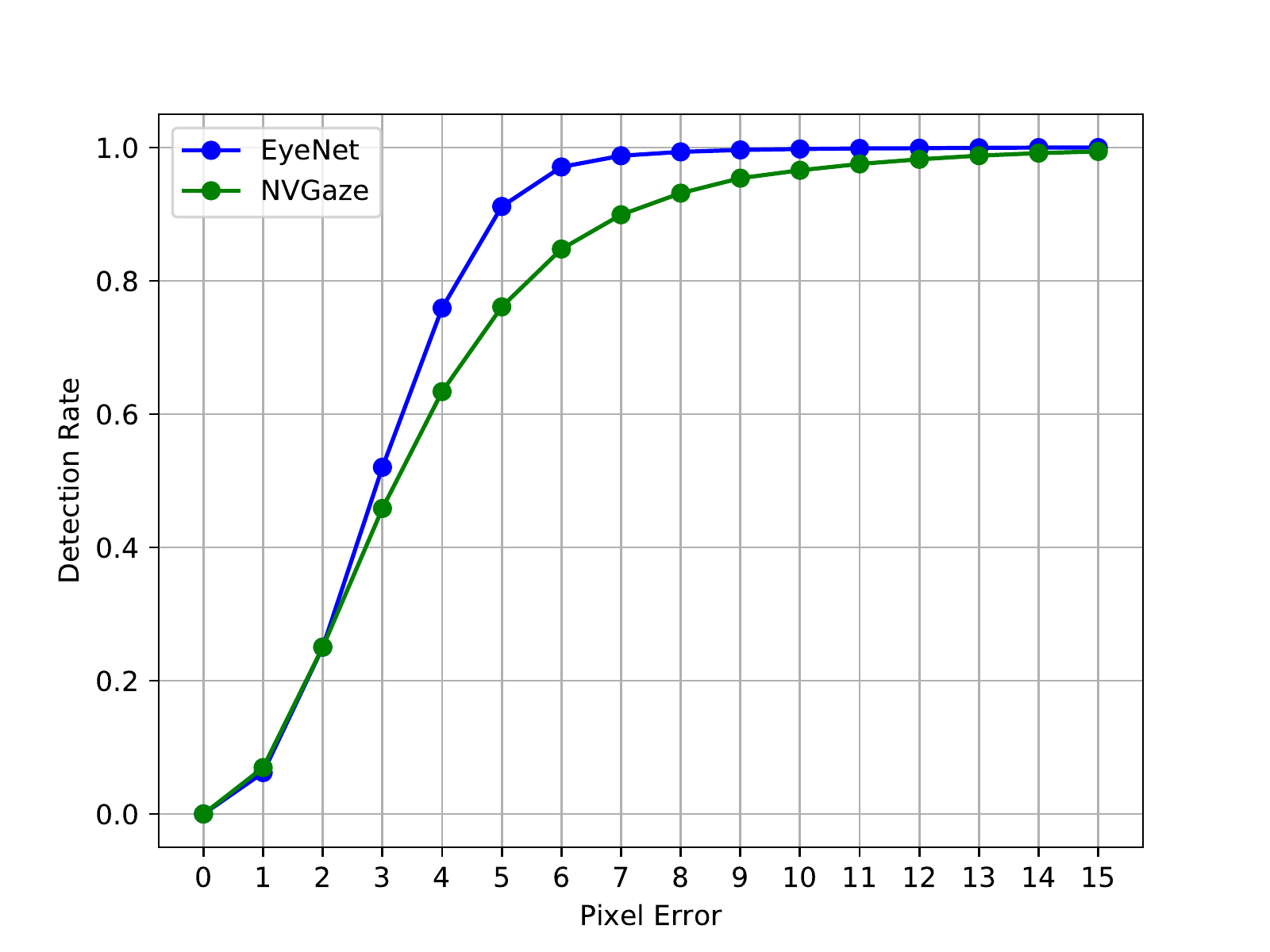} \\
        (a) & (b) \\
    \end{tabular}
    \caption[9pt]{(a) Qualitative segmentation results. Row A shows a challenging case where both methods performs well. Row B,C show a case where RlTnet and EyeNet has small defect respectively. (b) Detection rate for pupil localization. EyeNet reaches 100\% detection rate quicker than NVGaze}
  \label{fig:qualitative}
 \end{figure}

The segmentation ground truth of 334 subjects from device 1 are used as training set and 93 subjects from device 1 are used as testing set for this task comparison. We report the confusion matrix of both methods for all classes in Table~\ref{tab:segmentation} below. Both methods produced very accurate segmentations, which is vital for eye tracking pipelines because most methods use segmentation boundaries to deduce the 2D pupil center, an integral component to predict the final gaze. Figure~\ref{fig:qualitative}(a) shows some qualitative results for both methods.

\begin{table}[h]
\small
\centering
\label{tab:segmentation}
\setlength{\tabcolsep}{0.5em}
\def\arraystretch{1}
\caption{Confusion Matrix reporting Segmentation accuracy per class. In each cell, EyeNet / RlTNet results are reported; higher accuracies are marked in bold}
\begin{tabular}{|c|c|c|c|c|}
\hline
EyeNet/RITNet & Pupil          & Iris           & Sclera         & BG             \\ \hline
Pupil                  & 96.25 / \textbf{98.85} & 3.75 / \textbf{1.15}    & 0.00  / 0.00         & 0.00  / 0.0         \\ \hline
Iris                   & \textbf{0.04} / 0.42        & \textbf{99.03} / 97.85 & \textbf{0.93} / 1.47          & \textbf{0.00} / 0.26          \\ \hline
Sclera                 & 0.00 / 0.00          & 3.27 / \textbf{2.45}          & \textbf{96.71} / 93.77 & \textbf{0.02} / 3.78          \\ \hline
BG                     & 0.01 / \textbf{0.00}          & 0.72 / \textbf{0.04}          & 2.09 / \textbf{0.29}          & 97.18 / \textbf{99.67} \\ \hline
\end{tabular}
\vspace{-5mm}
\label{RIT}
\end{table}

\vspace{-3mm}
\subsection{Pupil Localization}
The pupil localization task is to predict the 2D pupil center location of the eye images. A precise pupil center estimate is crucial to deduce the 3D pupil location, which in turn is important to predict the final gaze.

For this task, a recent work titled NVGaze \cite{kim2019nvgaze} and EyeNet predictions are compared. Both works use end-to-end network to directly predict 2D pupil locations with a slightly different approach. NVGaze uses simple euclidean loss to directly estimate a 2D vector as pupil center while EyeNet uses a 2D probability map with cross-entropy loss to get the peak probability point as 2D pupil center. Since both methods downsample input images to a lower resolution for efficiency reasons, during our evaluation, we first resize dataset image to each method's specific input resolution, then scale back the predicted pupil to original full resolution to report the errors for a fair comparison.

Similar to segmentation, we use pupil center ground truth of 334 and 93 subjects in device 1 as training and testing set respectively. The pupil localization error is defined as the euclidean distance between predicted and labeled pupil center location. NVGaze produced an average pupil localization error of \textbf{4.88} while EyeNet produced average error of \textbf{4.22} on $640\times480$ resolution.

\vspace{-5mm}
\subsection{Gaze Estimation}
Gaze estimation involves predicting the direction of the visual axis. To parameterize the gaze direction as a pair of horizontal, vertical angles we need a nodal point or origin, for example the 3D cornea center. Given this origin, the vector from the origin to the central target direction on the virtual screen can be considered the reference direction from which relative directions to other targets can be measured and trained upon.

For synthetic datasets such as the one primarily used in NVGaze \cite{kim2019nvgaze} the origin is known (its the eye ball center). In real datasets such as MagicEyes we are forced to rely on an estimated origin to parameterize gaze directions for training and evaluation. To this end, we use the average position of the 3D cornea center (see Sec. \ref{sec:geometry} for cornea 3D center estimation) for the calibration targets as the origin during training and similarly the average position of the testing targets during evaluation. Note also that in real settings the estimated origin is also used to compensate for slippage of head mounted devices.

With the angular parameterization of gaze the objective for NVGaze is to learn the end-to-end mapping from image to gaze direction. In contrast, EyeNet uses its estimates of the cornea and pupil 3D center to estimate the optical axis and the DeepGazeMapper to turn this to the gaze vector.

Both methods are pre-trained using the 334 subjects of device 1 and then personalized on the calibration targets of the test data of device 2 to benchmark their cross-device generalization capability. Both methods are evaluated on the remaining targets of test set from device 2. Note that for EyeNet, we can perform online optimization/fine tuning of the DeepGazeMapper using the calibration targets. Therefore, for fair comparison with NVGaze we report the results with and without this optimization.



\begin{table*}[h]
\small
\centering
\setlength{\tabcolsep}{0.5em}
\caption{Gaze estimation on MagicEyes for EyeNet and NVGaze \cite{kim2019nvgaze}. EyeNet performs significantly better than the end-to-end NVGaze approach in terms of both accuracy and robustness}
\begin{tabular}{|c|c|c|c|c|c|c|c|c|}
\hline
Model& Description          & Mean AE      & Std AE                & Q1 AE    & Q2 AE       & Q3 AE             \\ \hline
1 & NVGaze & 285.67 & 203.00 & 110.71 & 233.30 & 483.42 \\ \hline
2 & EyeNet-DeepMapper  & 238.85 & \textbf{114.88} & 147.27 & 231.96 & 327.98  \\
\hline
3 & EyeNet-Opt-DeepMapper & \textbf{179.53} & 135.81 & \textbf{80.29} & \textbf{136.64} & \textbf{241.85} \\
\hline
\end{tabular}
\label{tab:gaze}
\end{table*}
The overall gaze estimation metric is defined as the angular
error between the true gaze vector (origin to gaze target) and the estimated gaze vector in arcmin units. Table~\ref{tab:gaze} shows all the statistics from both methods on gaze estimation. As we can see, EyeNet outperforms NVGaze in every metric, primarily because its use of a more explicit multi-step geometric model instead of an end-to-end approach. We believe the performance of NVGaze can be improved significantly by using more training targets covering a wide range of gaze. However, this is unsuitable in practice since it is very time consuming and tiresome.

During the training of NVGaze, its training error drops to as low as 80 average arcmin but the test loss cannot be dropped further beyond the results reported in Table~\ref{tab:gaze}. For test set, NVGaze produced predictions with high variance, so although it generates accurate results sometimes, it is not suitable for real world use. We added stronger regularization to combat overfitting but we were not able to produce significantly better results. We reason that incorporating pupil localization, 3D pupil and cornea estimation and optical-to-gaze calibration in the same network explicitly, will not generate precise enough results.

\vspace{-3mm}
\section{Conclusion}
In this work, we introduced the first large scale eye gaze estimation dataset, MagicEyes, collected with real mixed reality devices. After evaluating recent state-of-the-art algorithms on MagicEyes, we find, unsurprisingly, that fully convolutional networks performed well on semantic tasks such as eye segmentation and pupil localization. This bodes well for geometric methods which so far were plagued by robustness issues in segmentation, pupil detection in real world settings.  We also observed that the results of direct end-to-end deep learning methods for gaze estimation are still not competitive with hybrid learning and geometric approaches such as EyeNet. We hope that with the release of MagicEyes, the AR/MR eye tracking community will make rapid progress on building more accurate end-to-end learning based methods.

\appendix
\section{Appendix}
We present the details of model architectures we used in the experiments as well as additional visualization of MagicEyes images and segmentation results.
\subsection{EyeNet}
We attach the details of the EyeNet architectures in this section for reference.
\subsubsection{Encoder Architecture}
We emplot a ResNet50 with a feature pyramid network (FPN) as the encoder architecture of EyeNet~\cite{he2016deep},\cite{lin2017feature}. This part can be replaced with newer architectures for potential improvements. We select the pyramid layer with stride of 8 pixels with respect to the input image. Thus, the topmost layer encoder features have a size of 20x15x256. This shared feature is input to the task branches.

Note also that the original image capture size in our dataset is 640x480. However, we downsample it to \textbf{160x120} for computational efficiency as the input to the encoder.

\subsubsection{Decoder Architectures}
We tabulate each task decoder (branch) in the following. All the components are standard.

\begin{table*}[!htbp]
\centering
\begin{tabular}{|c|c|c|c|c|}
\hline
Layer Name     & Parent Layer   & Layer Params       & Input Size & Output Size \\ \hline
DeConv1        & Encoding Layer & 4x4x128(stride 2) & 20x15x256  & 40x30x128   \\ \hline
ResidualBlock1 & DeConv1        & 128                & 40x30x128  & 40x30x128   \\ \hline
ResidualBlock2 & ResidualBlock1 & 128                & 40x30x128  & 40x30x128   \\ \hline
DeConv2        & ResidualBlock2 & 4x4x64(stride 2)   & 40x30x128  & 80x60x64    \\ \hline
Conv1          & DeConv2        & 3x3x32             & 80x60x64   & 80x60x32    \\ \hline
BN1+ReLU       & Conv1          & NA                 & 80x60x32   & 80x60x32    \\ \hline
Conv2          & BN1            & 3x3x16             & 80x60x32   & 80x60x16    \\ \hline
Conv3          & Conv2          & 3x3x8              & 80x60x16   & 80x60x8     \\ \hline
DeConv3        & Conv3          & 4x4x4(stride 2)    & 80x60x8    & 160x120x4   \\ \hline
\end{tabular}
\caption{The architecture of the eye parts segmentation branch of EyeNet. See Section 3.2 for more details of this task.}
\end{table*}

\begin{table*}[!htbp]
\centering
\begin{tabular}{|c|c|c|c|c|}
\hline
\textbf{Layer Name} & \textbf{Parent Layer} & \textbf{Layer Params} & \textbf{Input Size} & \textbf{Output Size} \\ \hline
Conv1\_Loc          & DeConv2               & 3x3x32                & 80x60x64            & 80x60x32             \\ \hline
BN\_Loc +ReLU       & Conv1\_Loc            & NA                    & 80x60x32            & 80x60x32             \\ \hline
Conv2\_Loc          & BN\_Loc               & 3x3x20                & 80x60x32            & 80x60x20             \\ \hline
DeConv\_Loc         & Conv2\_Loc            & 4x4x5                 & 80x60x20            & 160x120x5            \\ \hline
\end{tabular}
\caption{The architecture of the pupil and glint localization branch of EyeNet. See Section 3.3 for more details of this task.}
\end{table*}


\begin{table*}[!htbp]
\centering
\begin{tabular}{|c|c|c|c|c|}
\hline
\textbf{Layer Name} & \textbf{Parent Layer} & \textbf{Layer Params} & \textbf{Input Size} & \textbf{Output Size} \\ \hline
Conv1\_Cor          & Encoding Layer             & 3x3x64                & 20x15x256           & 20x15x64             \\ \hline
BN1\_Cor+ReLU       & Conv1\_Cor            & NA                    & 20x15x64            & 20x15x64             \\ \hline
Conv2\_Cor          & BN1\_Cor              & 3x3x16                & 20x15x64            & 20x15x16             \\ \hline
BN2\_Cor+ReLU       & Conv2\_Cor            & NA                    & 20x15x16            & 20x15x16             \\ \hline
Flatten             & BN2\_Cor              & NA                    & 20x15x16            & 4800                 \\ \hline
FC1\_Cor            & Flatten               & 4800x64               & 4800                & 64                   \\ \hline
FC2\_Cor            & FC1\_Cor              & 64x16                 & 64                  & 16                   \\ \hline
FC\_Out\_Cor        & FC2\_Cor              & 16x2                  & 16                  & 2                    \\ \hline
\end{tabular}
\caption{Cornea 2D center estimation architecture. See Section 3.4 for more details of this task.}
\end{table*}

\subsection{RlTNet}
In order to maintain the exact same model architecture of the original RlTNet \cite{chaudhary2019ritnet}, we resize the dataset resolution 640x480 to the default resolution of RlTNet of 640x400 and re-scale it back to compare with the original ground truth. Though the aspect ratio change might have some negative effect on RlTNet, the much higher input resolution than EyeNet should compensate it to have a fair comparison.

\subsection{NVGaze}
Following the similar approach as for RlTNet, we also trade input dimension change for identical model architecture for NVGaze. Specifically, we downsample the original dataset image to 293x293 and 127x127 for pupil localization and gaze estimation respectively, according to the original paper \cite{kim2019nvgaze}.

\subsection{More Visualization of Dataset}
\begin{figure}[ht]
  \centering
    \includegraphics[width=\textwidth,keepaspectratio]{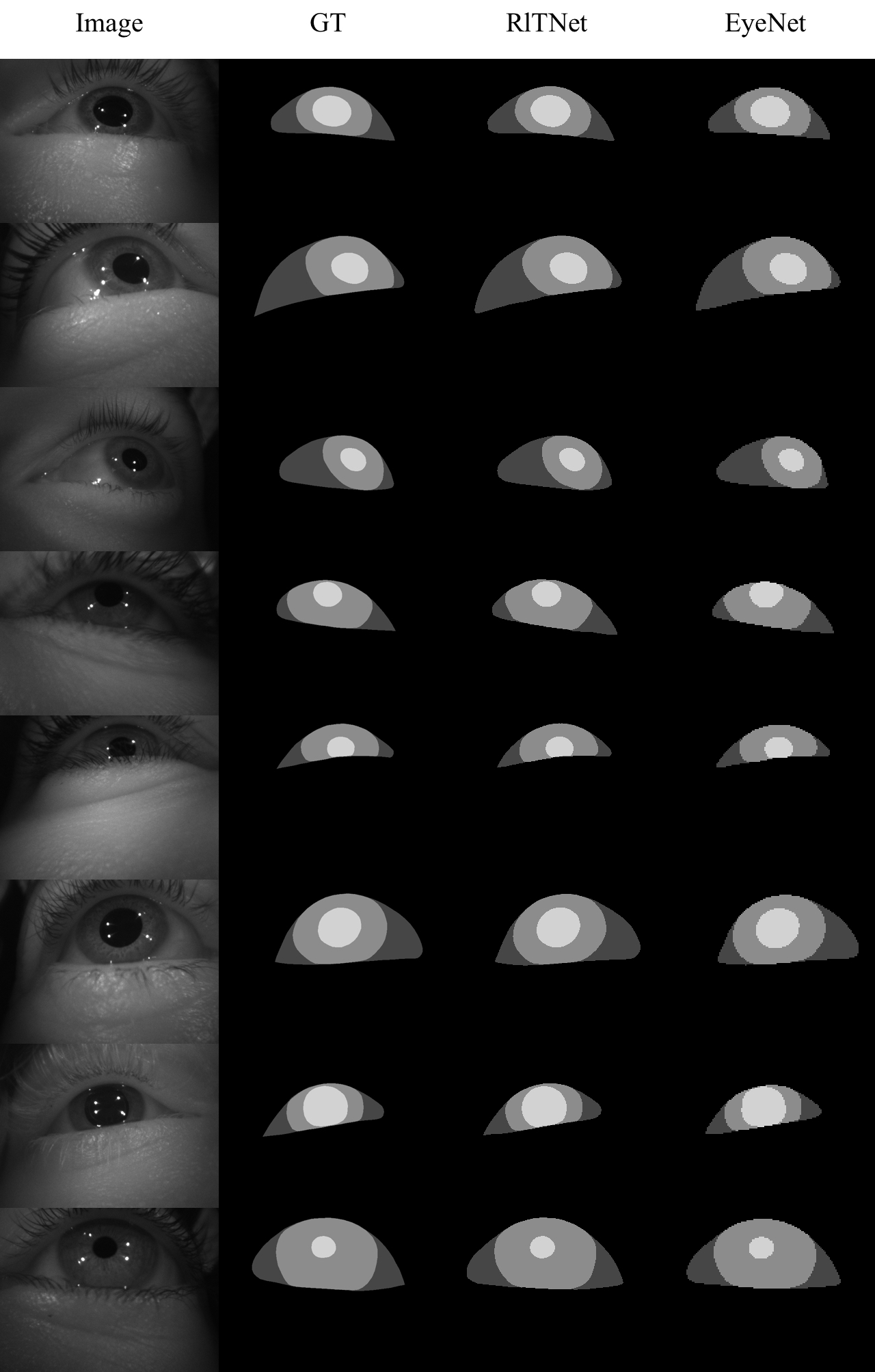}
  \label{fig:more_example}
\end{figure}

%
%
\clearpage
\bibliographystyle{splncs04}
\bibliography{magiceyes}
\end{document}